\documentclass{article}

\usepackage[nonatbib]{neurips_2021}

\usepackage[utf8]{inputenc} 
\usepackage[T1]{fontenc}    
\usepackage{hyperref}       
\usepackage{url}            
\usepackage{booktabs}       
\usepackage{amsfonts}       
\usepackage{nicefrac}       
\usepackage{microtype}      
\usepackage{xcolor}         
\usepackage{comment}
\usepackage{amsmath}
\usepackage{graphicx}
\usepackage{xspace}

\makeatletter
\DeclareRobustCommand\onedot{\futurelet\@let@token\@onedot}
\def\@onedot{\ifx\@let@token.\else.\null\fi\xspace}

\def\ie{\emph{i.e}\onedot}

\makeatother
\title{On the inductive biases of deep domain adaptation}

\author{%
  Rodrigue Siry\\
  UNICAEN - GREYC - CNRS\\
  \texttt{rodrigue.siry@unicaen.fr} \\
  \And
  Louis Hémadou\\
  ENPC\\
  \texttt{louis.hemadou@eleves.enpc.fr} \\
  \And
  Loïc Simon\\
  UNICAEN - ENSICAEN - GREYC - CNRS\\
  \texttt{loic.simon@ensicaen.fr} \\
  \And
  Frédéric Jurie\\
  SAFRAN\\
  \texttt{frederic.jurie@safrangroup.com} \\
}

\begin{document}

\maketitle
\begin{abstract}
Domain alignment is currently the most prevalent solution to unsupervised domain-adaptation tasks  and are often being presented as minimizers of some theoretical upper-bounds on risk in the target domain. However, further works revealed severe inadequacies between theory and practice: we consolidate this analysis and confirm that imposing domain invariance on features is neither necessary nor sufficient to obtain low target risk. We instead argue that successful deep domain adaptation rely largely on hidden inductive biases found in the common practice, such as model pre-training or design of encoder architecture. We perform various ablation experiments on popular benchmarks and our own synthetic transfers to illustrate their role in prototypical situations. To conclude our analysis, we propose to meta-learn parametric inductive biases to solve specific transfers and show their superior performance over handcrafted heuristics.

\end{abstract}
\section{Introduction}
Deep learning models achieve impressive performance on image classification tasks, when a large amount of training samples is provided. However, in many situations, abundant labeled data is not readily available for the task of interest (the target domain), while it exists in a related domain (the source domain) but with a statistical bias that makes it impossible to use as is to train an algorithm in the target domain.  The objective of {\em domain adaptation} is to exploit the data in the source dataset to obtain a model that performs well in the target domain.  In this paper, we focus on unsupervised domain adaptation, in which we only have access to unlabeled target samples during training.

Domain alignment (e.g., \cite{dann, associativeda, dsne, mdd}) is the dominant approach for solving unsupervised domain adaptation problems that fall under the covariate shift assumption, i.e., when the two domains contain the same classes and the same labeling rule. 
Their goal is to find a feature representation that is sufficiently informative to predict labels in the source domain while being domain independent in the distributional sense.
These methods are based on a series of theoretical results related to unsupervised domain adaptation: such as \cite{bendavid,mansour,SurveyDomainAdaptationTheory,mdd} which proposed various upper bounds on target risk. Indeed, in almost all these contributions, the bound includes - among other things - the source risk and some measure of deviation between source and target input data distributions.

However, further analyses of \cite{critique_alignement_1,critique_alignement_2,inductive_bias} have shown that practical domain-alignment algorithms are in fact largely inconsistent with the theory from which they claim to be derived and may even be counterproductive in some cases.  
Even if alternative methods \cite{da_self_supervision, da_disentangled} begin to emerge, domain-alignment remains prevalent in the literature and gives decent results on various transfers. This raises the question of why these methods work in some cases if not because of the invoked upper-bound rationale. 
This debate constitutes the central theme of this paper.
We argue that practical success of alignment can be better explained thanks to implicit inductive biases found in the standard practice of domain alignment literature.
In other words, such biases help to ensure that domain alignment behaves well despite the lack of conclusive theoretical guarantees. An inductive bias is a set of hidden assumptions that condition the behavior of the model on unseen data. Although \cite{inductive_bias} has already introduced the notion of inductive bias for domain alignment, this paper is to go further by giving a more comprehensive presentation of existing biases. Finally, as a complement to this expository work, we also propose a strategy to establish such inductive biases by meta-learning. Contrary to \cite{meta_da_1, meta_da_2} that employ meta-learning in the multi-source domain generalization setting (test on same classes but new unseen domain), we rather evaluate its ability to perform the same transfer with new test classes.

\section{Background and related work}
We call \textit{source} and \textit{target} domains two distributions $S$ and $T$ over the space of labeled data $X\times Y$. We define $S_X$ and $T_X$ the marginals of $S$ and $T$ over $X$ and $S_Y$, $T_Y$ the marginals over $Y$. Finally, we call \textit{domain shift} the discrepancy existing between $S$ and $T$. We study \textit{unsupervised domain adaptation}: provided labeled samples from $S$ and unlabeled samples from $T$, we would like to exploit labeled knowledge from $S$ to obtain a model that minimizes the target risk.
Obviously, this can only be possible if a relationship between $S$ and $T$ exists. The following general cases are often described in the literature:
i) {Covariate shift / inductive transfer}: $S_X \neq T_X$ and $P_S(Y | X) = P_T(Y | X)$,
ii) {Concept shift / transductive transfer}: $S_X = T_X$ and $P_S(Y | X) \neq P_T(Y | X)$,
iii) {Unsupervised transfer}: $S_X \neq T_X$ and $P_S(Y | X) \neq P_T(Y | X)$.

Most deep learning domain adaptation invoke the covariate shift assumption and several works build upon theoretical upper bounds on the target error. 
This is the case for example of \cite{dann} who built upon the upper bounds developped in \cite{bendavid}. 
Other works such as \cite{mdd} and \cite{WassersteinDAShen2017} develop their own bound and derive an alignment algorithm from it.
The upper bound usually takes the following form (see e.g. \cite{bendavid, mansour, WassersteinDAShen2017} or \cite{SurveyDomainAdaptationTheory} for a more exaustive list).
\begin{equation}
    \label{eq:gub}
    \begin{split}
        \epsilon_T(h) \leq& \epsilon_S(h) + \delta(S_X, T_X) + \lambda
    \end{split}
\end{equation}
where $h\in\mathcal{H}$ is a hypothesis, $\epsilon_S(h)$ and $\epsilon_T(h)$ represent the error associated to $h$ in both domains. 
The third term $\delta(S_X, T_X)$ represents a divergence between the marginal distribution of $X$ under both distributions, and this divergence involves the class of hypotheses $\mathcal{H}$.
The last term $\lambda$ underpins the adaptability in the sense that $\lambda$ is small only if a common hypothesis $h^*$ can perform well on both domains.
This term is not amenable to be computed without the labels in the target, but it is expected to be small in well posed problems.
In the remainder of this section, we recap a few facts that are not widely understood about current domain adaptation challenges and modern alignment approaches.

\paragraph{Fact 1: covariate shift is often meaningless.}
Indeed, in deep vision transfers, the two distributions have often disjoint supports ($S_X\perp T_X$). For instance, if the source domain is SVHN and the target is MNIST, no image belongs to both domain at the same time. In such cases, the notion of conditional distribution, say $P_S(Y|X)$, is undetermined outside of the support of $S_X$. As a result the constraint $P_S(Y|X)=P_T(Y|X)$ is nonbinding. 
Although this fact is often disregarded in domain adaptation theoretical works, which still advocate for the covariate shift assumption, practical takes on the problem introduce a feature extractor $\Psi:\mathcal X\to \mathcal Z$ hence embedding both distributions in latent space $\mathcal{Z}$ where the supports may overlap (e.g. \cite{coral, dan, dann, associativeda,mdd}).
Practical works go even further than making the supports overlap, they actually try to minimize a part of an upper-bound such as Eq~\ref{eq:gub}.
Since, the last term $\lambda$ cannot be evaluated without knowing the target labels, only the first two terms are considered in the minimization.
In essence, this boils down to searching for a feature extractor that is informative on labels in the source domain (first term), and insensitive to the domain (second term).
Therefore, the alignment {\em per se} is enforced through the minimization of the divergence term $\delta(S_X, T_X)$.

\begin{figure}
    \centering
    \hfill
    \includegraphics[scale=0.3]{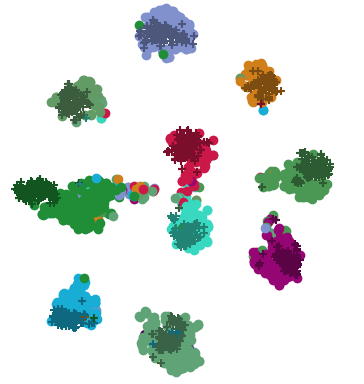}
    \hfill
    \includegraphics[scale=0.35]{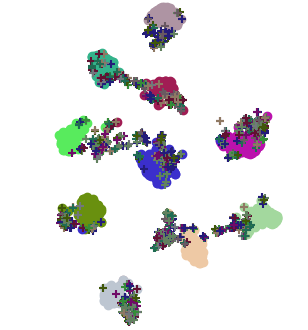}    
    \hfill\null
    \caption{t-SNE plot of feature space after DANN adaptation (labels are color-coded). Left: SVHN$\rightarrow$MNIST, Right: MNIST$\rightarrow$ SVHN.
    SVHN has many more degrees of freedom (digit fonts, scales, colors) than MNIST, it is therefore not surprising that features learnt from raw SVHN supervision transfer consistently on MNIST, making the joint feature distributions close. Left: one can see that the alignment obtained with DANN is non-degenerate.
    Right: The transpose transfer MNIST $\rightarrow$ SVHN is much harder. Features learnt with MNIST supervision do not extrapolate consistently to the richer SVHN. In that case, applying the DANN algorithm leads to a degenerate alignment. 
    }
    \label{fig:tsne}
\end{figure}

\paragraph{Fact 2: the feature extractor can be harmful.} This fact has been highlighted by at least three previous publications \cite{critique_alignement_1,critique_alignement_2,inductive_bias} in a form of a {\em no-free-lunch} theorem. 
The gist of this result relies on the observation that while the first two terms of the upper-bound are minimized, the third one can go out of control.
In particular, the feature extractor can map different regions of $T_X$ of inconsistent classes to the same region of $\mathcal Z$ leading to a larger target Bayes risk.
Worst, nothing prevents regions of $T_X$ of a given class to be embedded in regions corresponding to different classes of $S_X$.
This label mixing will be further strengthened by minimizing the divergence in scenarios where classes are balanced differently in the source and target domains (a situation known as {\em a prior shift} and  that is not ruled out by the covariate shift assumption).
Both phenomenons are due to the non-invertibility of $\Psi$ \cite{johansson2019support} (which is necessary to make the support of $S_X$ and $T_X$ overlap). 

\paragraph{Fact 3: domain alignment works well in many practical cases.} In practice, despite the previous caveat, domain alignment approaches based on upper-bounds are still successful in practice. 
On one hand, this success can be explained by a selection of published results. 
For instance, while the transfer from SVHN to MNIST is often reported, its transposed vestion (MNIST to SVHN) is never (see Fig~\ref{fig:tsne} for a visual comparison of DANN latent representation in both cases). 
On the other hand, the analysis of their success is restricted to reasoning on bounds such as Eq~\ref{eq:gub}, while the no-free-lunch theorem should raise concerns on these justifications. 
It is therefore clear, that other explanations should be put forward to understand what makes a transfer amenable to a favorable alignment.
In this article, we elaborate on the role of (hidden) inductive biases, as advocated by \cite{inductive_bias}.

\section{The role  of (implicit) inductive bias in domain adaptation}\label{sec:biases}

Given that theoretical bounds alone cannot enlighten entirely what makes domain adaptation work, we propose here a set of experiments to complement the understanding of several cases of inductive bias.
We organize the different biases in four main categories corresponding to: i) biases involved in the domain alignment {\em per se}, ii) data augmentation , iii) implicit bias of the feature extractor and iv) network architecture.
All experiments involve training a classifier of the form $f = g\circ \Psi$, where $\Psi$ is a feature extractor and $g$ is a classifier. 

The experiments presented in this section are based on different datasets commonly used in domain adaptation, \ie  MNIST \cite{mnist}, MNIST-M \cite{dann}, SVHN \cite{netzer2011reading} and Office-31 \cite{office31}. Moreover, we use the default ResNet-18 encoder for $\Psi$ in all experiments except for the BatchNorm ablation one (in that case, a simpler VGG-16 convolutional architecture is used). When applicable, we use the default ImageNet pre-trained weights provided by PyTorch. The classifier $g$ is a simple feedforward network with one hidden layer of width 1024.

\subsection{Inductive biases in the alignment method}
Our first experiment is designed to evaluate the response of several domain alignment strategies with respect to prior shift. 
To introduce such a shift while keeping the experiment compatible with the covariate shift assumption, we consider two versions of MNIST for both domains. 
In the source domain the classes are represented in a balanced ratio (i.e. $10\%$), while in the target we purposely make the ratios uneven, with class partitions ranging from $5$ to $20\%$.
As a result, the transfer problem falls strictly speaking within the covariate shift assumption, \ie $P_S(Y|X)=P_T(Y|X)$.

In this experiment we consider three forms of domain alignment. The first one, corresponds to training $g\circ \Psi$ without any explicit alignment. In that case, referred to as {\em source only} (SO), the cost function is computed solely from the cross entropy on source samples. The second and third approaches correspond to DANN \cite{dann} and associative DA \cite{associativeda}.
As expected, in case of prior shift, trying to align marginal distributions in the latent representation $Z=\Psi(X)$ is detrimental. 
This can be seen in Table~\ref{tab:equilibrate}. Indeed, on one hand, the source-only training performs almost perfectly ($0.99$ target accuracy).
On the other hand, the DANN approach tries to strictly impose the alignment of marginal and deteriorates the performance by a large margin ($0.67$ target accuracy). 
Last, the associative DA approaches the alignment in a looser manner, by relying on a {\em cluster assumption} and reaches a target accuracy of $0.97$.
In fine, even though the alignment remains slightly harmful to the performance, this experiment exhibits a form of inductive bias that renders alignment more robust to prior shift.

 \begin{table}[ht]
    \centering
    \begin{tabular}{c c c c}
       \toprule
        & SO & DANN & ASSO-DA\\
       \midrule
        M$\rightarrow$ MI& 0.99 & 0.67 & 0.97\\
       \bottomrule
    \end{tabular}
    \caption{Effect of alignment in a prior-shifted transfer; M= balanced MNIST, MI=MNIST with class imbalance}
    \label{tab:equilibrate}
\end{table}

\subsection{Data Augmentation}
A straightforward and well-known way to enhance feature generalization is to explicitly augment input images with a family of synthetic, class-preserving image transforms. For example, it can be constructed by a composition of random flips, scaling, rotation, translation, blur or changes in color statistics. Such transforms represent a subset of all possible class-preserving transforms and can hence partly explain domain shifts found in real life. Therefore, augmentation of source images is another inductive bias whom effect on transferability must be measured in the domain adaptation setting. In figure \ref{tab:augment}, we show that a simple color randomization of source MNIST digits and backgrounds helps solving the hard MNIST$\rightarrow$SVHN transfer, usually out of reach even for pre-trained models, with a drastic $45\%$ increase of target performance. This shows how a simple, but ad-hoc transform can help bridging the complex domain shift existing between MNIST and SVHN.

 \begin{table}[ht]
    \centering

     \begin{tabular}{ccc}
       \toprule
        & DANN w/o augment & DANN w/ augment\\
       \midrule
       M$\rightarrow$ S & 0.15& 0.6\\
       \bottomrule
    \end{tabular}
    \caption{Effect of augmentation; M=MNIST, S=SVHN; Model is pre-trained}
    \label{tab:augment}
\end{table}

\subsection{Pretraining and optimizer implicit bias}

The next question is probably the most subtle one. It has been noted by \cite{siry2020study} that the target accuracy of a source-only trained network is predictive of high chance of success in the alignment itself.
This observation can lead to several interpretations in terms of inductive bias.
We envision the following two possibilities (the latter being a refined version of the former): i) pretraining the feature extractor leads to an operating point where source and target samples are clustered in a consistent way in the latent embedding or ii) pretraining the feature extractor and the optimization bias of the source cross-entropy leads to such a consistently clustered embedding. 
To evaluate the likelihood of these two potential explanations, we have conducted two experiments as described hereafter.

 \begin{table}
    \centering
    \begin{tabular}{c c c c}
       \toprule
       Transfer & NoPT+FT & PT & PT+FT\\
       \midrule
       M$\rightarrow$ MM & 0.13 & 0.19 & 0.56\\

       MM$\rightarrow$ M & 0.98& 0.29 & 0.97\\

       S$\rightarrow$ M &0.56& 0.25 & 0.84\\

       M$\rightarrow$ S &0.06& 0.19 & 0.23\\

       A$\rightarrow$ W &0.26& 0.91 & 0.88\\

       D$\rightarrow$ W &0.49& 1.0 & 0.99\\

       D$\rightarrow$ A &0.14& 0.77 & 0.75\\

       W$\rightarrow$ A &0.19& 0.71 & 0.69\\
       \bottomrule
    \end{tabular}
    \caption{Target accuracies of the KNN classifier on static pre-trained features (PT) and features obtained after source-only fine-tuning (PT+FT). We also give non-pretrained source-only features as a reference (NoPT+FT); A=Office31-Amazon, W=Office31-Webcam, D=Office31-Dslr, M=MNIST, MM=MNIST-M, S=SVHN}
    \label{tab:knn}
\end{table}

\paragraph{Assessing feature transferability with KNN:}
First, we design a simple KNN classifier to predict the label of some feature sample from the target distribution.
It simply finds the top-K (K=50) source nearest neighbors of the unlabeled target query sample and returns the majority vote of their K labels.
Good accuracy of this classifier hence proves natural class equivariance and domain invariance of the considered feature space.
However, as mentioned earlier, expecting the property to be verified for frozen initial ImageNet features might be too restrictive since the source-only training baseline also involves end-to-end tuning of the encoder model on source samples, which subsequently modifies the feature space.
To account for this alternate explanation, the effects of the source-only fine-tuning must be reported as well.
Therefore, we evaluate our KNN classifier on ImageNet pre-trained ResNet-18 features, before (PT) and after (PT+FT) source-only fine-tuning. We also evaluate non-pretrained features obtained after source-only (NoPT+FT) as a baseline.
We do so for several transfers and report results in Table~\ref{tab:knn}.
On digit classification transfers, we observe better-than-random but low KNN accuracy with frozen features, for example: 19\% for MNIST$\rightarrow$ MNIST-M and 25\% for SVHN$\rightarrow$ MNIST. However, source and target feature distributions become significantly closer as the encoder is tuned on source, with KNN accuracies rising to 56\% and 84\% respectively on those two transfers. On Office-31 transfers, pre-trained features display good KNN accuracy both before and after being tuned. This might be due to the relative similarity existing between Office-31 and ImageNet images, suppressing the need to further specialize the features.
The benefits of pre-training hence cannot be fully explained by the geometry of pre-trained output features: pre-trained hidden ResNet-18 layers also condition the dynamics of source-only fine tuning that eventually leads to good source and target representations thanks to an implicit optimization bias.

\paragraph{Dead Pixel CIFAR:}
To complement this first experiment on the influence of pretraining, we have designed an experiment to show that the presence of confusing factors in the source domain can make pretraining detrimental. 
Note that this specific experiment is purposely extreme: in fact, it is extreme to the point of making a commonly useful bias such as ImageNet pretraining fail completely.
Indeed, we build a synthetic transfer tailored to make pre-training perform worse than random initialization.
In both domains, we construct an image sample by taking a random CIFAR-10 image and set its $i^{th}$ upper-left pixel to a fixed color value (grey), the pixel index $i$ is chosen between 1 and 10.
In both domains, the task is to evaluate $i$ which means that the class is tied to the so-called Dead Pixel.
In the source domain, $i$ matches the original CIFAR-10 image class.
On the contrary, in the target domain, the dead pixel index and the original image class are completely decorrelated, and the image label is given by the dead pixel and not by the object contained in the image.

This experiment, however artificial, exposes one of the main obstacles in domain adaptation, which consists in confusion factors that are present in the source domain but absent from the target one.
When supervised on source, we expect pretraining to induce a bias associated with the CIFAR image to label correlation.
It is therefore highly probable that a pre-trained model shall ignore the dead pixel and exploit features from the CIFAR-10 object, as the initial parameters already make this information salient and filter out pixel-level details.
In the target domain, where the CIFAR-10 object is not useful anymore to predict the image label, the pre-trained model should fail even if source-only fine-tuning is performed.
On the contrary, a randomly initialized classifier should quickly identify the dead pixel as the safest, easiest way of predicting the image label and should completely ignore the CIFAR-10 object. Consequently, the classifier trained from scratch on source should achieve a good accuracy on target.

In Table~\ref{tab:pre_train}, we summarize results of source-only fine-tuning and DANN on several standard transfers as well as the Dead Pixel CIFAR transfer. 
We observe a consistent performance improvement on most transfers when pre-training is used in both source-only and DANN adaptation. While DANN provides an additional increase in performance on most digit transfers ($+46\%$ on MNIST $\rightarrow$ MNIST-M,  $+6\%$ on SVHN$\rightarrow$MNIST without pre-training), it has on average a negative impact on Office-31 performance when pre-training is used (ex: decrease of $8\%$ in DSLR$\rightarrow$Webcam), putting into question its ability to really improve on a representation that already displays good properties of domain-invariance and class equivariance. In the Dead-Pixel-CIFAR transfer, pre-training performs worse but still manages to exploit some Dead-Pixel information, leading to reasonable performance on target. The model trained from scratch performs consistently in both domains and reaches maximum accuracy on target.
What is more, explicit alignment with DANN does hardly improve over source only in this scenario.

In this section, we have confirmed that pre-trained representations are usually responsible for reaching features displaying a high degree of transferability.
We have shown that pretraining alone is not always sufficient and is mostly useful when combined with source-only fine-tuning.
Our synthetic experiment corroborates that pre-training should be considered as an inductive bias that helps with domain shifts similar to those found in real life, and fails in the general case where confusing factors may make this bias counter-effective.

 \begin{table}
    \centering
    \hfill
    \begin{tabular}{ccc}
       \toprule
        & SO & DANN\\
       \midrule
       M$\rightarrow$ MM &0.17& 0.63\\

       MM$\rightarrow$ M & 0.98& 0.98\\

       S$\rightarrow$ M & 0.65& 0.71\\

       M$\rightarrow$ S & 0.07& 0.1\\

       A$\rightarrow$ W & 0.27& 0.33\\

       D$\rightarrow$ W & 0.51& 0.72\\

       D$\rightarrow$ A & 0.15& 0.16\\

       W$\rightarrow$ A & 0.18& 0.13\\
       D-Pix &0.95&0.99\\
       \bottomrule
    \end{tabular}
    \hfill
     \begin{tabular}{ccc}
       \toprule
        & SO & DANN\\
       \midrule
       M$\rightarrow$ MM & 0.6& 0.99\\

       MM$\rightarrow$ M & 0.98& 0.98\\

       S$\rightarrow$ M & 0.83& 0.88\\

       M$\rightarrow$ S & 0.25& 0.15\\

       A$\rightarrow$ W & 0.88& 0.84\\

       D$\rightarrow$ W & 1.0& 0.92\\

       D$\rightarrow$ A & 0.76& 0.74\\

       W$\rightarrow$ A & 0.71& 0.79\\
       D-Pix & 0.75 & 0.76\\
       \bottomrule
    \end{tabular}
    \hfill\null
    \caption{Target accuracies for the source-only baseline and the DANN alignment algorithm; \textbf{Left}: No pretraining, \textbf{Right}: pretraining; A=Office31-Amazon, W=Office31-Webcam, D=Office31-Dslr, M=MNIST, MM=MNIST-M, S=SVHN, D-Pix=Dead Pixel CIFAR transfer}
    \label{tab:pre_train}
\end{table}

 \subsection{Architectural network components}\label{subsec:ib_architecture}
 The update of features during source-only tuning is a complex, non-linear process that does not depend solely on the initialization of the feature encoder.
 To have a finer understanding of how transferable task-relevant features emerge, one must also take into account the architecture of the encoder.
In this section, we conduct additional experiments to show how transfer learning can be sensitive to these choices.
 
\paragraph{The BatchNorm} is mainly used to accelerate training of deep models: it rescales all activations in the effective range of the subsequent non-linearity, avoiding flat regions in the loss landscape on which gradient descent makes little progress. Furthermore, this multiplicative operation might give rise to descriptors that display higher degrees of invariance to certain transformations. We perform a very simple transfer, whose purpose is to adapt MNIST to its color-inverted counterpart, Inv-MNIST, and show in Table~\ref{tab:batch_norm} that in the non-pretrained case, adding BatchNorm entirely conditions success of domain adaptation for both source-only and DANN.
 
 \begin{table}[ht]
    \centering
    \begin{tabular}{c c c c}
       \toprule
       M$\rightarrow$ InvM & SO & DANN\\
       \midrule
       Encoder-BN & 0.5 & 0.95\\

       Encoder-NoBN & 0.05 & 0.04\\
       \bottomrule
    \end{tabular}
    \caption{Source only accuracies for two different encoder architectures (with and without batch normalization); M=MNIST, InvM=MNIST with inverted colors }
    \label{tab:batch_norm}
\end{table}

\paragraph{Global Pooling} is an operation that collapses a whole feature map into a single vector by averaging across all spatial dimensions. The mean operator does not retain the original location of features and is therefore translation-invariant. Popular backbone architectures VGG-16, ResNet-18 or DenseNet have a 7x7 AvgPool layer. We conduct another transfer to evaluate the capacity to learn translation-invariance without being explicitly supervised to do so: the source domain is MNIST, while the target domain is MNIST-T. MNIST-T is generated from MNIST samples, augmented by a random translation/scale/rotation. We report results of this transfer in Table~\ref{tab:pool}. Results show again how a simple, but appropriate architectural inductive bias such as Average Pooling can modify the extrapolation behavior of a model on an unseen target domain. 
Interestingly, when this positive bias is used, DANN brings further improvement over SO, while without global pooling, DANN performs even worth than SO.

\begin{table}
    \centering
    \begin{tabular}{c c c c}
       \toprule
       M$\rightarrow$ MT & SO & DANN\\
       \midrule
      NoGlobalAvgPool & 0.51 & 0.48\\

       GlobalAvgPool & 0.8 & 0.95\\
       \bottomrule
    \end{tabular}
    \caption{Source only accuracies for two different encoder architectures; M=MNIST, MT=MNIST with random translation/scale/rotation augmentation }
    \label{tab:pool}
\end{table}
 \paragraph{Dropout} \cite{dropout} is a simple yet very effective technique to regularize the capacity of neural networks by setting activations to zero with some probability $p$ during training. This avoids co-adaptation of neurons and performs implicit model averaging. As dropout is ubiquitous in standard classification to increase generalization, we would like to measure how it contributes to domain-invariance. To do so, we run all experiments of Table~\ref{tab:pre_train} again with and without dropout in the penultimate layer.
 We've noticed that the impact of dropout was not the same depending whether pre-training is applied as well or not. We therefore gather in Table~\ref{tab:drop_out} a detailed sets of results (with/without dropout and with/without pretraining).
 Comparison with the non-dropout baseline shows a slight but consistent improvement of target performance in the non-pretrained case and negligible impact in the pre-trained one. In transfers that are out of reach even for pre-trained models (MNIST$\rightarrow$ SVHN), dropout is helpless.

\begin{table}
\centering
\hfill
    \begin{tabular}{ccc}
       \toprule
        & SO & DANN\\
       \midrule
       M$\rightarrow$ MM &0.17 / 0.13& 0.63 / 0.7\\

       MM$\rightarrow$ M & 0.98 / 0.98& 0.98 / 0.985\\

       S$\rightarrow$ M & 0.65 / 0.65&  0.71 / 0.74\\

       M$\rightarrow$ S & 0.07 / 0.07& 0.1 / 0.1\\

       A$\rightarrow$ W & 0.27 / 0.28 & 0.33 / 0.35\\

       D$\rightarrow$ W & 0.51 / 0.6& 0.72 / 0.79\\

       D$\rightarrow$ A & 0.15 / 0.2& 0.16 / 0.17\\

       W$\rightarrow$ A & 0.18 / 0.2& 0.13 / 0.17\\
       \bottomrule
    \end{tabular}
    \hfill
     \begin{tabular}{ccc}
       \toprule
        & SO & DANN\\
       \midrule
       M$\rightarrow$ MM & 0.6 / 0.7& 0.99 / 0.98\\

       MM$\rightarrow$ M & 0.98 / 0.98& 0.98 / 0.984\\

       S$\rightarrow$ M & 0.83 / 0.8& 0.88 / 0.91\\

       M$\rightarrow$ S & 0.25 / 0.22& 0.15 / 0.14\\

       A$\rightarrow$ W & 0.88 / 0.88& 0.84 / 0.84\\

       D$\rightarrow$ W & 1.0 / 1.0& 0.92 / 0.91\\

       D$\rightarrow$ A & 0.76 / 0.75& 0.74 / 0.75\\

       W$\rightarrow$ A & 0.71 / 0.73& 0.79 / 0.75\\
       \bottomrule
    \end{tabular}
\hfill\null
\caption{Target accuracies without / with dropout; \textbf{Left table:} Non-pre-trained \textbf{Right table:} Pre-trained; A=Office31-Amazon, W=Office31-Webcam, D=Office31-Dslr, M=MNIST, MM=MNIST-M, S=SVHN}
    \label{tab:drop_out}
\end{table}

\section{Meta Learning of Inductive Biases}\label{sec:metalearning}
Until now, we have been looking for inductive biases that are either heuristic or handcrafted. Meta-learning, or "learning to learn", is a recent trend in machine learning that aims at learning the inductive bias from data itself, by directly optimizing the purpose it should serve. In this section, we propose to meta-learn a parametric inductive bias that helps better transfer some given domain shift.\\

\textbf{General definition of meta-learning:} A meta learning model aims at solving any task $\mathcal{T}$ sampled from some distribution $p(\mathcal{T})$, each task is composed of its own training and test data. Tasks from $p(\mathcal{T})$ are assumed to share some common information. To this matter, the model parameter is split in two subsets $\theta$ and $\Phi$:
\begin{itemize}
    \item $\theta$ is the task-specific parameter, or "fast weight": to learn a specific task $\mathcal{T}_i$ presented on-the-fly, it is allowed to be modified into $\theta_i$ during an imposed heuristic called "inner-loop training". Its initial value also follows a fixed distribution.
    \item $\Phi$ is the task-agnostic parameter, or "meta-parameter", or "slow weight": it embeds the common information shared among all $\mathcal{T} \sim p(\mathcal{T})$. It is not allowed to change when a specific $\mathcal{T}_i$ is provided, but is meta-optimized to a fixed value that maximizes expected performance of $(\theta_i, \Phi)$ on every $\mathcal{T}_i$. This process is called "meta-optimization" or "outer-loop training".
\end{itemize}

Any prototype of meta-learning model can then be defined by some parametrization of $\theta$, $\Phi$ and some inner-loop design. We expect a clever meta-learning design to beat on average any heuristic optimization/regularization on $\mathcal{T}_i$ once meta-trained.
Meta-training $\Phi$ requires related third-party data: the usual practice is to split the dataset of tasks in two subsets meta-train and meta-test, and hope that the value $\Phi$ optimized for meta-train tasks will also perform well on meta-test tasks.

\textbf{Application to domain adaptation:}
In the previous part, we concluded that a good inductive bias for domain adaptation should be able to condition the model to be invariant to all visual transforms related to the domain shift while being equivariant to class-related information. 

We hence propose to meta-learn an initialization $\Phi$ for a classifier that maximizes its source-only performance on a given transfer $S\rightarrow T$: after $t$ SGD steps on training batches from $S$, $\Phi_t$ should give a good predictor for any sample from $T$. This model is a simple variant of MAML on which train and test data are domain-shifted and that is not constrained to the few-shot regime. To catch up with our formalism, note that here $\theta_i = \Phi_t - \Phi$. While this training loop might be similar to \cite{meta_da_1}, we do not evaluate the meta-model in the context of multi-source domain generalization (same classes, unseen domain) but measure its ability to solve the same transfer for tasks involving new, unseen classes.

In our experiments, we use images from the VisDA dataset, composed of $6$ domains of various difficulty containing 345 classes each. We always use the "Real" images domain as source, and meta-train $\Phi_{Real\rightarrow T}$ for any remaining target domain $T$. We define a task as a 10-way source-only training problem. We meta-train on the first 200 classes and test on the remaining ones. To reduce model size and training time, we work with frozen ResNet-18 pre-encoded features in all baselines. 

\textbf{Baselines:}
We compare our meta-model to several baseline models to assess superiority of parametric inductive biases.
\textbf{Random initialization:} A randomly initialized classifier, directly tuned on the downstream 10-way test task. \textbf{Pre-train and Fine-tune:} Random initialization might not be a fair baseline, as it could not exploit labeled information from the 200 meta-train classes from both domains. We hence propose to pre-train the classifier model to solve the 200-way classification problem for both meta-train domains simultaneously, then replace the last layer and fine-tune on the downstream 10-way test task. \textbf{DANN:} We perform the 10-way transfer with both labeled source images and unlabeled target images, we add the DANN adversarial constraint in the classifier to enhance transferability. \textbf{Pre-train and DANN:} We cumulate the pre-trained initialization and DANN training. Note that none of these transfers are prior-shifted as we use artificially balanced source and target batches, and that in all cases, we favor the baselines by choosing the best optimizer, learning rate and number of fine-tune iterations.

We report results in Table~\ref{tab:maml}. In all cases, each transfer-specialized MAML (MAML-2DOM) beats all baselines by a significant margin, especially on the real $\rightarrow$ quickdraw transfer on which ResNet features are not amenable to transferability, hence requiring correction by an appropriate inductive bias in the downstream classifier. Heuristic alignment such as DANN only improves slightly over Random. In both source-only and DANN fine-tuning, pre-training on meta classes does not improve over random initialization.

\begin{table}
    \centering
    \begin{tabular}{cccccc}
       \toprule
        & Random+SO & Pretrain+SO & DANN & Pretrain+DANN & MAML-2DOM\\
       \midrule
       real$\rightarrow$quickdraw &0.185\scriptsize{ $\pm$ 0.036} & 0.197\scriptsize{ $\pm$ 0.053}& 0.191\scriptsize{ $\pm$ 0.033} & 0.300\scriptsize{ $\pm$ 0.082} & \textbf{0.542}\scriptsize{ $\pm$ 0.010}\\

       real$\rightarrow$painting &0.708\scriptsize{ $\pm$ 0.065} & 0.641\scriptsize{ $\pm$ 0.065}& 0.731\scriptsize{ $\pm$ 0.067} & 0.728\scriptsize{ $\pm$ 0.050} & \textbf{0.767}\scriptsize{ $\pm$ 0.009}\\

       real$\rightarrow$sketch &0.501\scriptsize{ $\pm$ 0.053} & 0.447\scriptsize{ $\pm$ 0.043}& 0.536\scriptsize{ $\pm$ 0.054} & 0.576\scriptsize{ $\pm$ 0.082} & \textbf{0.681}\scriptsize{ $\pm$ 0.015}\\

       real$\rightarrow$clipart & 0.620\scriptsize{ $\pm$ 0.059} & 0.543\scriptsize{ $\pm$ 0.061}& 0.656\scriptsize{ $\pm$ 0.082} & 0.640\scriptsize{ $\pm$ 0.071} & \textbf{0.746}\scriptsize{ $\pm$ 0.008}\\

       real$\rightarrow$infograph & 0.359\scriptsize{ $\pm$ 0.086} & 0.313\scriptsize{ $\pm$ 0.032}& 0.446\scriptsize{ $\pm$ 0.067} & 0.415\scriptsize{ $\pm$ 0.062} & \textbf{0.502}\scriptsize{ $\pm$ 0.005}\\
       \bottomrule
    \end{tabular}
    \caption{Average target accuracies and standard deviations for 10-way domain adaptation test tasks, computed over 10 runs; we outline best performance in bold}
    \label{tab:maml}
\end{table}
\section{Conclusion}

In this paper, we pursued the analysis initiated by \cite{critique_alignement_1, critique_alignement_2}. 
As a starting point, we consider the following question.
How much of the success of deep domain alignment approaches can be unraveled through theoretical upper-bounds from domain adaptation theory.
Despite their appealing prospects and their prescriptive significance in terms of modern approaches, a no-free-lunch theorem can be stated which invalidates a universal benefit of the domain alignment strategy.
This surprising fact calls for a refined analysis of the key ingredients of successful domain alignment transfers.

We therefore investigate the role of various inductive biases to give a more appropriate account of the situation.
We illustrate four kinds of inductive biases ranging from those inherent to the alignment approach itself to more generic ones such as the network architecture or data augmentation.
In particular, the role of pretraining on a general purpose database such as imagenet is insidious for several reasons. 
First, without this step, current alignment methods often fail which suggests that pre-training helps moving the supports of source and target distributions to a close match from which alignment methods can catch up.
It appears nonetheless that the role of pretraining is sometimes more tangled as was evidenced by our kNN classifier experiment.
In a nutshell, in such situations it is the combined effect of pretraining and source-driven optimization biases that are responsible for successful alignment.

The latter evidence motivated our search for efficient ways to design good inductive biases in a principled way. 
We have conducted an illustrative experiment in which we meta-learned parametric inductive biases that perform better than usual domain-adaptation heuristics on a given transfer. 
Given the impact of pre-training indicated by our former analysis, we proposed to meta-learned the initialization of the network.
Although this choice revealed very effective in our set of experiments, it is not the only meaningful avenue. 
For instance, it is also possible to approach meta-learning on a regularization perspective or an optimization one. We hope to see more future work exploring this direction. 
On a closing note, we would like to point out that if our analysis sheds some light on the inner working of domain alignment, it is only in an empirical way: much remains to be done on the theoretical counterpart.

\bibliographystyle{IEEEtran}
\bibliography{References}

\end{document}